# Label Confidence Weighted Learning for Target-level Sentence Simplification

**Xinying Qiu** and **Jingshen Zhang**

Department of Computer Science, School of Information Science and Technology
Guangdong University of Foreign Studies, China

xy.qiu@foxmail.com

## Abstract

Multi-level sentence simplification generates simplified sentences with varying language proficiency levels. We propose Label Confidence Weighted Learning (LCWL), a novel approach that incorporates a label confidence weighting scheme in the training loss of the encoder-decoder model, setting it apart from existing confidence-weighting methods primarily designed for classification. Experimentation on English grade-level simplification dataset shows that LCWL outperforms state-of-the-art unsupervised baselines. Fine-tuning the LCWL model on in-domain data and combining with Symmetric Cross Entropy (SCE) consistently delivers better simplifications compared to strong supervised methods. Our results highlight the effectiveness of label confidence weighting techniques for text simplification tasks with encoder-decoder architectures.

## 1 Introduction

Text simplification aims to reduce the linguistic complexity of a text while preserving its meaning, making it more accessible to a wider audience, such as language learners, individuals with cognitive impairments, or those with low literacy skills. Multi-level sentence simplification takes this a step further by generating simplified sentences tailored to specific target audiences with varying language proficiency levels.

Despite the importance of multi-level sentence simplification, progress in this area has been hindered by the limited availability of labeled parallel corpora. The only available dataset for this task, Newsela-auto (Xu et al., 2015, Jiang et al., 2020), is relatively small, making it challenging to train high-performing models. To address this issue, previous studies have explored data augmentation techniques, such as constructing pseudo-parallel corpora using complex-to-simple sentence pairs (Nishihara et al., 2019, Katsuta and Yamamoto, 2019). However, these approaches may suffer from the propagation of label errors in the augmented corpora, leading to suboptimal performance.

In this paper, we propose Label Confidence Weighted Learning (LCWL), a novel approach to multi-level sentence simplification that leverages weak supervision from a large-scale paraphrase dataset and a pre-trained classifier to generate pseudo-labeled simplification data. It mitigates the impact of noisy labels through a label confidence weighting scheme in the training loss of the encoder-decoder model. This sets LCWL apart from existing confidence-weighting methods primarily designed for classification tasks. We evaluate our approach on the Newsela-auto dataset and demonstrate that LCWL outperforms state-of-the-art unsupervised baselines across multiple complexity levels. Fine-tuning the LCWL model on in-domain data further improves performance, consistently delivering superior simplifications compared to strong supervised methods. Our results highlight the effectiveness of label confidence weighting techniques for learning with noisy pseudo-labels in text simplification tasks involving encoder-decoder architectures. We provide our codes at: https://github.com/astro-jon/LCWL.

## 2 Related Work

### 2.1 Text Simplification and Target-level Simplification

Text simplification aims to reduce the linguistic complexity of a text while preserving its meaning, making it more accessible to a wider audience (Siddharthan 2014a). Early approaches focused on lexical and syntactic simplification using rule-based methods (Devlin and Tait, 1998), while recent data-driven approaches treat simplification as a monolingual machine translation task, learning from complex-simple parallel corpora (Xu et al. 2015, Nisioi et al. 2017, Martin et al. 2020a, Sun et al. 2023). Kriz et al. (2020) proposed Simple-QE, a BERT-based model for estimating the quality of simplified texts that correlates well with human judgements.

Target-level simplification extends this idea by generating output tailored to specific readability levels or reader profiles. Scarton and Specia (2018) introduced the concept and developed a model that uses artificial tokens to control simplification complexity. Subsequent work expanded on this idea by considering factors like reading ease scores (Nishihara et al. 2019), simplification operations (Agrawal et al. 2021), and multi-tasking (Chi et al. 2023). Kew and Ebling (2022) explored target-level simplification by training models on corpora with automatically labeled complexity levels.

However, a key challenge in target-level simplification is the lack of large-scale parallel corpora annotated with target complexity levels. Existing resources like Newsela (Xu et al. 2015) are relatively small, hindering the development of high-performing models. Our work addresses this issue by proposing Label Confidence Weighted Learning (LCWL), which leverages weak supervision from a large-scale paraphrase dataset and a pre-trained classifier to generate pseudo-labeled simplification examples across various complexity levels.

### 2.2 Learning from Noisy Labels

Training accurate deep neural networks (DNNs) in the presence of label noise is a critical challenge, as DNNs can easily overfit to noisy labels and suffer from poor generalization (Arbit et al. 2017, Zhang et al. 2021, Song et al. 2022). Existing methods for learning with noisy labels can be categorized into five groups: robust architecture, robust regularization, robust loss function, loss adjustment, and sample selection (Song et al. 2022).

Loss adjustment methods, particularly loss reweighting, are closely related to our proposed LCWL approach. These methods modify the training loss to reduce the impact of noisy examples by assigning different weights to them. Importance reweighting (Wang et al. 2017) assigns weights based on the ratio of clean and noisy data distributions, which can be challenging to estimate accurately. Active Bias (Change et al. 2017) emphasizes uncertain examples with inconsistent predictions, but may incorrectly focus on genuinely ambiguous examples. DualGraph (Zhang et al. 2021) uses graph neural networks to reweight examples based on label relations, but requires a specific architecture and may not capture instance-dependent noise.

Robust loss functions, such as Symmetric Cross Entropy (SCE) (Wang et al. 2019), introduce additional terms to make the learning process more noise-tolerant. However, these approaches have been primarily designed and evaluated for classification tasks, and their adaptation to encoder-decoder models for sequence-to-sequence tasks remains an open question. Our proposed methodology is designed to optimize the cross-entropy loss, similar to SCE by Wang et al. (2019), but operated on the decoder module instead of the classifier. We choose SCE as our comparison baseline for confidence weighting method and also explore the combination of both approaches.

## 3 Methodology

We propose Label Confidence Weighted Learning (LCWL), a novel approach that leverages weak supervision from a large-scale paraphrase dataset and a pre-trained classifier to generate pseudo-labeled simplification examples across various complexity levels. LCWL combines loss reweighting and sample selection techniques, allowing the model to mitigate the impact of noisy labels and exploit additional unlabeled data.

### 3.1 Label Confidence Weighted Learning

To address the challenge of limited high-quality supervised data for multi-level text simplification, we propose a novel approach that leverages paraphrase data and label confidence weighted learning. Our methodology consists of the following steps as shown in Figure 1.

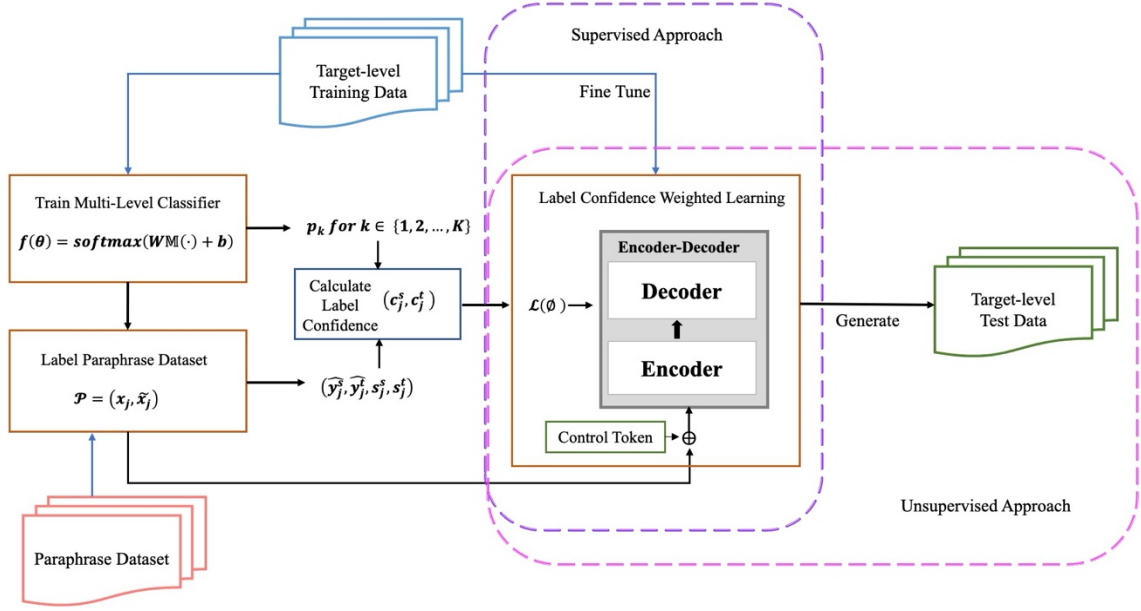

Figure 1 Research Structure with Label Confidence Weighted Learning

**(1) Train a multi-level simplification classifier:**

We train a classifier $f_\theta(x) \rightarrow y$ on a labeled target-level benchmark dataset $\mathcal{D} = (x_i, y_i)_{i=1}^N$ (i.e. Newsela-auto training set) for sentence simplification to predict the simplification level $y \in 1, 2, ..., K$ of a given sentence $x$.

**(2) Label the paraphrase dataset:**

We apply the trained classifier $f_\theta$ to both the source and target sentences of a large paraphrase dataset $\mathcal{P} = \{(x_j, \tilde{x}_j)\}_{j=1}^M$. We generate pseudo training data $\mathcal{D}' = \left(x_j, \tilde{x}_j, \widehat{y_j^s}, \widehat{y_j^t}, s_j^s, s_j^t\right)_{j=1}^M$ for sentence simplification. For each sentence pair $(x_j, \tilde{x}_j)$, we record the predicted simplification levels $\widehat{y_j^s}$ and $\widehat{y_j^t}$ and the classifier's confidence scores $s_j^s$ and $s_j^t$ for the source and target sentences, respectively.

**(3) Train an encoder-decoder model with label confidence weighting:**

We use the pseudo-parallel data $\mathcal{D}'$ to train an encoder-decoder model $g_\phi(x, y^s) \rightarrow \tilde{x}$ for text simplification, where $y^s$ is the predicted simplification level of the source sentence. To mitigate the impact of mislabeled sentences, we introduce a label confidence weighting scheme in the training loss:

**a.** Calculate the precision $p_k$ of the simplification classifier $f_\theta$ for each level $k \in \{1, 2, ..., K\}$.

**b.** For each labeled sentence pair $\left(x_j, \tilde{x}_j, \widehat{y_j^s}, \widehat{y_j^t}, s_j^s, s_j^t\right)$, compute label confidence scores $c_j^s = \sqrt{p_{\widehat{y_j^s}} \cdot s_j^s}$ and $c_j^t = \sqrt{p_{\widehat{y_j^t}} \cdot s_j^t}$ for the source and target sentences, respectively, as the geometric mean of the classifier's precision for the predicted level and the sentence's confidence score.

**c.** During training, minimize the label confidence weighted cross-entropy loss:

$$\mathcal{L}(\phi) = -\frac{1}{M} \sum_{j=1}^M c_j^s \cdot c_j^t \cdot \log p\left(g_\phi(x_j, \widehat{y_j^s}) = \tilde{x}_j\right)$$

By incorporating label confidence weights $c_j^s$ and $c_j^t$ into the training loss, we aim to reduce the influence of mislabeled examples and improve the model's ability to generate appropriate simplifications.

### 3.2 BERT-based Multi-level Classifier

For the multi-level classifier, we use BERT (Devlin et al. 2018) which has been experimentally proven effective (see Appendix B

for details), extract the last hidden layer representation of the [CLS] token ($h_b$) for sentence $x_i$ and use it to predict the sentence label, where $W$ and $b$ are parameters.

$$h_b = \text{BERT}(x_i)$$
$$f_\theta = \text{softmax}(Wh_b + b)$$

### 3.3 BART-based Generation Model

Let $x = (x_1, x_2, ..., x_n)$ be the input sentence and $y = (y_1, y_2, ..., y_m)$ be the target simplified sentence at a specific level. We prepend to each input sentence a single special token indicating the target level such as <SIMP_3> for level 3. The BART model (Lewis et al. 2020) can be written as:

$$h^{enc} = Encoder(x)$$
$$P(y|x, l) = Decoder(h^{enc}, l)$$

where $h^{enc}$ is the encoded representation of the input $x$, and $l$ is the target simplified level. The training objective is to minimize the label confidence weighted cross-entropy loss $\mathcal{L}(\phi)$. During inference, the model generates the simplified sentence $y$ autoregressively using the encoder representation $h^{enc}$ and the target simplified level $l$.

## 4 Experiment Design

### 4.1 Dataset

**Newsela-auto**. We chose Newsela-auto, the only available multi-level text simplification corpus, with 1,912 English news articles professionally simplified to multiple grade levels (Jiang et al., 2020). We signed agreements with Newsela and the authors of Jiang et al. (2020) to use the dataset for research purposes only and will not redistribute it. Following Maddela et al. (2021), we filtered the pseudo-aligned data by retaining sentence pairs with pairwise BLEU scores between 0.1 and 0.9 to remove noisy alignments. The final grade-level distribution of Newsela-auto is shown in Table 1.

|          | Simp-1 | Simp-2 | Simp-3 | Simp-4 |
|----------|--------|--------|--------|--------|
| Train    | 669    | 768    | 595    | 354    |
| Validate | 108    | 109    | 76     | 17     |
| Test     | 182    | 233    | 199    | 133    |

Table 1: Newsela-auto graded-level distribution after filtering pairwise BLEU score

[1] https://www.cs.cmu.edu/~jwieting/

**ParaNMT**. We randomly selected 1 million sentence pairs from the processed version of the Para-NMT-50M dataset[1] (Wieting and Gimpel, 2018) as our paraphrase dataset for pseudo labeling. To ensure data quality, we filtered out poor alignments based on the following criteria: sequences that are identical, sequences contained within one another, sequences lacking letters, and sequences with fewer than 3 words.

### 4.2 Baseline Models

**(1) Unsupervised Approch:**

**MUSS** (Kew and Ebling, 2022) is a text simplification method that fine-tunes BART-large on a large paraphrase data set using ACCESS (Martin et al., 2020 and 2021) control tokens to manage target-specific output through a parameter search on the validation split.

**FUDGE-Target** (Kew and Ebling, 2022) adapts FUDGE (Yang and Klein, 2021) for text simplification by training a separate binary classifier for each simplification level. The classifiers are combined with a BART-large generator fine-tuned on paraphrase sentence pairs for target-level simplification.

**SCE** We apply Symmetric Cross Entropy (SCE) (Wang et al., 2019) while training the classifier using the Newsela-auto training set. By combining cross-entropy loss with a reverse cross-entropy term, SCE improves robustness to label noise. The SCE-optimized classifier is then used to label the paraphrase dataset without LCWL, enabling a direct comparison between SCE and our LCWL method as alternative confidence-weighting algorithms.

**LCWL** Our proposed methodology trains the generation model with label confidence weighted cross-entropy loss using only pseudo training data.

**LCWL+SCE**: We apply SCE (Wang et al., 2019) to train the multi-level classifier. After obtaining the pseudo-labels, we apply our proposed LCWL when training the encoder-decoder module, leveraging the benefits of both techniques for improving the simplification model's performance.

**(2) Supervised Approach:**

**SUPER** (Kew and Ebling, 2022) is a level-aware supervised baseline model following Scarton and Specia (2018). It prepends to each source sentence a single control token indicating the target level.

**GPT-3.5-Turbo** To our knowledge, no study has evaluated the capability of large language models (LLMs) for target-level sentence simplification, despite recent findings that LLMs outperform the best sentence simplification methods (Feng et al. 2023). We take the initiative to assess GPT-3.5-turbo's performance on target-level SS using a few-shot supervised approach. Our experiments involve prompting the LLM with 3 example pairs of original-target sentences for each of the 4 simplification levels (see Appendix E for the few-shot prompt template).

**SCE+FT**: We apply SCE during classifier training, use the optimized classifier to label the paraphrase dataset, and fine-tune the BART generator on the pseudo-labeled data. The model is then further fine-tuned using the Newsela-auto training set before being applied to the test set for target-level simplification, investigating the benefits of combining SCE with fine-tuning for supervised target-level SS.

**LCWL+FT** This model further fine-tunes the LCWL model on the Newsela-auto training set with a BART-based generation model before evaluating on the Newsela-auto test set.

**SCE+LCWL+FT**: We first apply SCE to optimize the multi-level classifier and label the paraphrase dataset. We then employ LCWL during the encoder-decoder training, considering the confidence of assigned labels. Finally, we fine-tune the generator using the Newsela-auto training set before applying it to the test set, aiming to maximize performance by combining SCE, LCWL, and fine-tuning.

## 5 Evaluation Metrics

We use seven metrics, $\Delta SLE$, SARI, LENS RF, LENS NoRF, $FKGL$, BERTScore, BLEU.

$\Delta SLE$ (Cripwell et al., 2023) ΔSLE, based on the reference-less Simplicity Level Estimate (SLE) metric, measures the relative simplicity gain of a system output compared to the input sentence by calculating the difference in their predicted simplicity levels.

**SARI** (Xv et al., 2016) evaluates sentence simplification by comparing system output with the original sentence and references, averaging F1 scores for three editing operations.

***FKGL*** (Flesch-Kincaid Grade Level) (Kincaid et al., 1975) measures readability gain based on word and sentence length.

**LENS** (Maddela et al. 2023): LENS is a learnable metric that evaluates text simplification quality by encoding the input, system output, and human references using Transformer models. Trained on human ratings with an adaptive loss focusing on the most relevant references, it predicts a quality score.

**LENS-SALSA** (Heineman et al. 2023): LENS-SALSA is a reference-free automatic simplification metric, trained to predict sentence- and word-level quality simultaneously.

**BERTScore** (Zhang et al., 2019) assesses semantic preservation between system output and references, but does not reflect simplification degree.

**BLEU** (Papineni et al., 2002) calculates the similarity between system output and references based on n-gram matching, regardless of position.

To comprehensively evaluate the performance of different models, we calculate their average rank across 7 metrics and 4 simplification levels. However, prior research has highlighted the potential drawbacks of using the BLEU metric for evaluating sentence simplification (Sulem et al., 2018), and BERTScore, while effective in assessing semantic similarity, falls short in quantifying the degree of simplification achieved. Therefore, we also report the average ranks of each model based on a subset of 5 evaluation metrics, excluding BLEU and BERTScore, to provide a more focused assessment of simplification quality.

## 6 Results and Analysis

**Unsupervised Approach:**

Tables 2 and 3 summarizes the performance of the unsupervised methods across various evaluation metrics. LCWL consistently outperformed other approaches, achieving first place in all evaluation metrics except for FKGL.

| | Evaluation Metrics | ΔSLE | LENS↑ | LENS-SALSA | SARI↑ | FKGL | BS↑ | BLEU↑ | Avg. Rank↓ |
|---|---|---|---|---|---|---|---|---|---|
| | **Simp-1** | 0.67 | | 70.697 | - | 9.66 | - | - | |
| Unsupervised Methods | MUSS | 0.69* | **63.28** | **70.46** | 35.69 | 7.75 | 75.95 | 41.29 | 3.15‖**2.4** |
| | FUDGE | 0.32 | 61.13* | 66.73 | 36.1 | **8.81** | 80.45 | **51.98** | 2.86‖3.2 |
| | SCE | 0.479 | 59.28 | 68.88 | 37.06* | 11.45* | 78.1 | 41.8 | 3.43‖3.2 |
| | *LCWL* | 1.73 | 60.86 | 69.43* | **37.78** | 7.1 | 87.41* | 43.07 | 2.86‖3 |
| | *LCWL+SCE* | **0.671** | 59.68 | 69.05 | 37.03 | 11.92 | **88.11** | 43.74* | **2.57**‖3 |
| Supervised Methods | SUPER | 0.07 | 65.03* | 66.981 | 32.5 | **9.36** | 88.2 | **75.06** | 3.3‖3.6 |
| | GPT-3.5-Turbo | **0.75** | 64.76 | 72.13 | **38.45** | 10.56 | 86.29 | 36.27 | 3‖**2.2** |
| | SCE+FT | 0.28* | 61.59 | **69.77** | 37.53* | 10.57 | 88.75* | 58.94* | **2.7**‖3 |
| | *SCE+LCWL+FT* | 0.277 | **65.43** | 67.91 | 36.8 | 10.51 | 88.3 | 58.8 | 3‖3 |
| | *LCWL +FT* | 0.19 | 63.19 | 68.28* | 37.1 | 10.29* | **90.08** | 57.3 | 3‖3.2 |
| | **Simp-2** | 1.3 | | 73.557 | - | 7.48 | - | - | |
| Unsupervised Methods | MUSS | 0.77 | 60.27* | **71.3** | 36.57 | 7.27* | 65.91 | 17.23 | 3.29‖2.6 |
| | FUDGE | 0.51 | 58.19 | 67.08 | **38.32** | 7.42 | 70.75 | 36.89 | 3.57‖3.4 |
| | SCE | 0.595 | 58.91 | 69.68 | 37.33 | 10.58 | **89.61** | 37.46 | 3.43‖4 |
| | *LCWL* | 1.74 | **61.84** | 70.78* | 38.27* | 7.14 | 89.54* | 38.15* | **1.86**‖**1.8** |
| | *LCWL+SCE* | 0.79* | 59.99 | 70.36 | 37.75 | 10.83 | 87.16 | **38.71** | 2.86‖3.2 |
| Supervised Methods | SUPER | 0.14 | 62.2 | 66.7 | 31.1 | 8.88 | 78.2 | **56.65** | 4.3‖4.8 |
| | GPT-3.5-Turbo | **0.88** | **67.16** | **74.02** | 41.62* | 9.58 | 87.8 | 33.4 | 2.7‖**2** |
| | SCE+FT | 0.61 | 63.8 | 72.04* | 39.4 | **7.82** | **96.92** | 52.3 | 2.7‖3 |
| | *SCE+LCWL+FT* | 0.73 | 65.26* | 71 | **42.7** | 8.35 | **96.92** | 55.58* | **2.3**‖2.6 |
| | *LCWL +FT* | 0.8* | 64.7 | 71.9 | 41.6 | 8.3* | **96.92** | 48.4 | 2.6‖2.6 |
| | **Simp-3** | 2.0 | | 75.669 | - | 5.88 | - | - | |
| Unsupervised Methods | MUSS | 1.49* | 57.02 | 71.3* | 38.05 | 5.19* | 56.03 | 10.55 | 3.43‖2.8 |
| | FUDGE | 0.81 | 52.69 | 68.23 | **39.56** | **6.44** | 61.46 | 23.98 | 3.43‖3.2 |
| | SCE | 0.65 | 58.17 | 70.68 | 37.51 | 10 | **89.61** | 33.19 | 3.57‖4.2 |
| | *LCWL* | 1.63 | **61.68** | **71.83** | 38.68* | 7.39 | 89.54* | 33.31* | **1.71**‖**1.6** |
| | *LCWL+SCE* | 0.83 | 59.51* | 71.08 | 38.22 | 10.13 | 87.16 | **34.09** | 2.86‖3.2 |
| Supervised Methods | SUPER | 0.66 | 61 | 66.5 | 37.9 | 6.65 | 66.6 | 39.6 | 4.4‖4.6 |
| | GPT-3.5-Turbo | 0.97 | 66.2 | 74.4 | 41 | 8.81 | 87.79* | 31.3 | 4‖4.2 |
| | SCE+FT | 1.59 | 66.64* | 74.4 | 41.7 | 5.4 | 82.9 | 40.29* | 2.9‖3 |
| | *SCE+LCWL+FT* | 1.81* | **67.98** | 75.13 | **46.14*** | 6.1* | **92.15** | **46.48** | **1.4**‖**1.6** |
| | *LCWL +FT* | **2.19** | 64.8 | **76.85*** | 47.11 | 5.87 | 74.7 | 34 | 2.3‖**1.6** |
| | **Simp-4** | 2.63 | | 77.639 | - | 4.16 | - | - | |
| Unsupervised Methods | MUSS | 1.41* | 55.23 | 71.15 | **39.63** | 5.61* | 51.73 | 7.65 | 3.14‖2.6 |
| | FUDGE | 1.04 | 41.64 | 61.69 | 37.03 | **4.6** | 49.6 | 11.06 | 3.86‖3.6 |
| | SCE | 0.69 | 58.94 | 71.16 | 35.18 | 8.81 | **87.77** | 26.9 | 3.43‖4 |
| | *LCWL* | 1.76 | **60.46** | **72.14** | 37.49* | 5.65 | 83.72* | 27.48 | **1.57**‖**1.6** |
| | *LCWL+SCE* | 0.852 | 59.89* | 71.71* | 37.32 | 9.32 | 82.85 | 27.07* | 3‖3.2 |
| Supervised Methods | SUPER | 1.53 | 58.9 | 62.6 | 43.2 | 5.09 | 55 | 24.5 | 4.3‖4 |
| | GPT-3.5-Turbo | 1.14 | 65.6 | 75.1 | 40.9 | 7.87 | **79.97** | 28.6 | 3.9‖4.6 |
| | SCE+FT | 1.98 | **67.16** | 74.7 | 42.1 | 3.95* | 63.7 | 31.47* | 2.9‖2.8 |
| | *SCE+LCWL+FT* | 2.33* | 62.9 | 76.52* | **46.42** | 4.01 | 65 | **39.27** | **1.9**‖**1.8** |
| | *LCWL +FT* | **2.69** | 65.42* | **77.52** | 46.23* | 4.73 | 75.51* | 26.4 | 2.1‖**1.8** |

Table 2: Comparison of unsupervised and supervised methods on Newsela-auto across 4 simplification levels using 7 evaluation metrics. ↑ indicates higher scores are better. For ΔSLE, LENS-SALSA, and FKGL, scores closer to the ground truth are better. The best and second-best performances are bolded and starred, respectively. Our proposed methodologies are bolded and italicized. Average ranks within supervised and unsupervised categories are calculated across 7 metrics and 5 metrics (excluding BERTScore and BLUE), separated by ‖ with the 7-metric average on the left, with highest ranks bolded.

| Avg. Rank ↓ over 4 Levels | 7-Metric | 5-Metric | ΔSLE | LENS | LENS-SALSA | SARI | FKGL |
|---|---|---|---|---|---|---|---|
| MUSS | 3.25 | 2.6* | 2.25* | 2.75* | 2* | 3.75 | 2.25* |
| FUDGE | 3.43 | 3.35 | 4 | 4.25 | 5 | 2.5* | **1** |
| SCE | 3.46 | 3.85 | 4.25 | 3.75 | 3.75 | 4 | 3.5 |
| LCWL | **2** | **2** | **2** | **1.5** | **1.5** | **1.75** | 3.25 |
| SCE+LCWL | 2.82* | 3.15 | 2.5 | 2.75* | 2.75 | 3 | 4.75 |

Table 3: Comparison of average ranks of **unsupervised methods** across 4 simplification levels using 7 evaluation metrics, 5 metrics excluding BERTScore and BLEU, and individual ranks on the 5 metrics separately. The best and second-best ranks are bolded and starred, respectively.

| Avg. Rank ↓ over 4 Levels | 7-Metric | 5-Metric | ΔSLE | LENS | LENS-SALSA | SARI | FKGL |
|---|---|---|---|---|---|---|---|
| SUPER | 4.07 | 4.25 | 4.75 | 3.75 | 5 | 4.5 | 3.25 |
| GPT-3.5-Turbo | 3.39 | 3.25 | 2.75 | 3.5 | **2.25** | 3 | 4.75 |
| SCE+FT | 2.79 | 2.95 | 3 | 3* | 2.75* | 3.25 | 2.75 |
| SCE+LCWL+FT | **2.14** | **2.25** | 2.5* | **1.75** | 2.75* | **2** | 2.25* |
| LCWL+FT | 2.5* | 2.3* | **2** | 3* | 2.25 | 2.25* | **2** |

Table 4: Comparison of average ranks of **supervised methods** across 4 simplification levels using 7 evaluation metrics, 5 metrics excluding BERTScore and BLEU, and individual ranks on the 5 metrics separately. The best and second-best ranks are bolded and starred, respectively.

| Avg. Rank ↓ over 4 Levels | 7-Metric | 5-Metric | ΔSLE | LENS | LENS-SALSA | SARI | FKGL |
|---|---|---|---|---|---|---|---|
| SCE+LCWL+FT | **1.68** | 1.8 | 2 | **1.25** | 2 | **1.75** | 2 |
| LCWL+FT | 1.71 | **1.6** | **1.5** | 1.75 | **1.5** | **1.75** | **1.5** |
| LCWL | 2.57 | 2.6 | 2.5 | 3 | 2.5 | 2.5 | 2.5 |

Table 5: Comparison of average ranks of **two best supervised methods** (SCE+LCWL+FT and LCWL+FT) and **the best unsupervised method** (LCWL). The best ranks are bolded.

LCWL's high scores in ΔSLE indicate significant simplification effectiveness, and its consistently high scores in SARI reflect its ability to balance simplification with content preservation.

MUSS performed well, achieving second place in most evaluations as shown in Table 3, but fell short in SARI. This suggests that while MUSS can maintain readability, as evidenced by its strong performance in LENS and LENS-SALSA metrics, it may not always achieve the desired level of simplification. FUDGE, in contrast, achieves the first place on FKGL and second on SARI but was behind the other methods on other metrics. This indicates that FUDGE excels at improving readability and balancing simplification with content preservation, but may not be as effective as LCWL in overall simplification performance.

SCE was underperforming compared to the others and was not improving LCWL when they are combined. This might be due to the limitations of SCE's confidence-weighting algorithm, which was originally designed for classification tasks. Additionally, the combination of SCE and LCWL may not provide significant improvements over LCWL alone because LCWL's label confidence weighting scheme is already effective at mitigating the impact of noisy labels for unsupervised approach. In conclusion, LCWL's outstanding performance across multiple evaluation metrics highlights the effectiveness of its label confidence weighting scheme in generating high-quality simplifications.

**Supervised Approach**

Tables 2 and 4 compare the performance of supervised methods. SCE+LCWL+FT achieves the best performances on 7-metric and 5-metric average, LENS and SARI ranking, and second best all others. LCWL+FT closely follows, achieving top ranks in ΔSLE, LENS-SALSA and FKGL, and second best on all others. While SCE's algorithm may not be as effective as LCWL's label confidence weighting scheme in the unsupervised approach, the combination of SCE and LCWL in the supervised approach (SCE+LCWL+FT) leverages their complementary strengths. SCE helps to reduce the impact of noisy labels during the classifier training phase, while LCWL's label confidence weighting scheme mitigates the impact of noisy pseudo-labels during the generator fine-tuning phase.

GPT-3.5-Turbo and SCE+FT displayed competitive performance, particularly in metrics like LENS and LENS-SALSA However, they were generally outperformed by LCWL+FT and SCE+LCWL+FT.

**Comparisons of Best Methods**

Table 5 provides a comparative analysis of the best methods from both the supervised and unsupervised approaches. LCWL+FT and SCE+LCWL+FT, the best supervised methods, consistently outperformed LCWL, the best unsupervised method, across various metrics.

The combination of LCWL's label confidence weighting with fine-tuning on in-domain data significantly boosted performance, setting a new benchmark for target-level sentence simplification. The unsupervised approach relies solely on the pseudo-labeled paraphrase dataset, which may not capture all the nuances of human-like simplifications. In contrast, fine-tuning step allows the model to adapt to the specific characteristics of the target dataset and learn to generate simplifications that are more aligned with human judgments.

The combination of label confidence weighting and fine-tuning emerges as a particularly powerful strategy, consistently delivering high-quality sentence simplifications across various complexity levels. This demonstrates the robustness and effectiveness of our proposed Label Confidence Weighted Learning approach in enhancing the performance of target-level sentence simplification models. We further provide case studies to compare the simplification results of various methods in Appendix D.

## 7 Case Analysis

Given the Case Studies in Appendix D, we examine the simplification results across different levels and reveal several key insights into the performance of our proposed methods and baselines:

**Sentence Splitting and Information Preservation**:

A notable pattern across simplification levels is the tendency of LCWL+SCE+FT and LCWL to split complex sentences into multiple simpler ones, particularly at higher simplification levels. This strategy significantly improves readability while preserving essential information.

For example, in Simplification Level-2, Case-1:

*Original*: "The scientists studied 22 very different species of finned swimmers, using video recordings, lab studies and computer modeling to determine what pattern, if any, might exist."

*LCWL+SCE+FT*: "The scientists studied 22 very different species of finned swimmers. They used video recordings, lab studies and computer modeling to determine what pattern, if any, might exist."

This splitting approach aligns well with human-written references and effectively simplifies the content.

**Contextual Information and Explanations:**

LCWL+FT and LCWL+SCE+FT often excel at providing additional context or explanations for complex concepts or unfamiliar terms, particularly in lower simplification levels. This helps maintain a balance between simplification and informativeness.

For instance, in Simplification Level-1, Case-2:

*Original*: "...including House Speaker John Boehner, R-Ohio, and Senate Minority Leader Mitch McConnell, R-Ky."

*LCWL+FT*: "...including House Speaker John Boehner, a Republican, and Senate Minority Leader Mitch McConnell, a Kentucky senator."

This addition of contextual information helps readers understand the roles and affiliations of the mentioned individuals without prior knowledge.

**Vocabulary Simplification**:

Across all levels, LCWL and its variants demonstrate a strong ability to replace complex vocabulary with simpler alternatives. This is particularly noticeable in higher simplification levels.

For example, in Simplification Level-3, Case-2:

*Original*: "While most of us hope to dodge space rocks, NASA has unveiled an ambitious, $105 million plan to build a spaceship to drag one closer to Earth."

*LCWL*: "most of us hope to avoid the space rocks, but NASA is planning to build a spaceship to bring one closer to earth."

The replacement of "dodge" with "avoid" and "drag" with "bring" makes the sentence more accessible to lower-level readers.

**Comparison with Baseline**

LCWL-based methods consistently outperform other approaches, such as SCE+FT, in terms of balancing simplification with content preservation. SCE+FT manages to split the sentences (Level-1 Cases 1 and 2, Level-3 Cases 1 and 2). But it sometimes oversimplifies by omitting crucial information (Level 2 Cases 1 and 2), and in many cases performs little vocabulary simplification or simply repeats the original sentence especially at higher simplification levels.

**Areas for Improvement:**

Despite the overall strong performance, there are some areas where the models could be improved:

*Consistency in splitting decisions*: While sentence splitting is generally effective, there are instances where LCWL doesn't split sentences when it would be beneficial.

*Information retention*: In some cases, particularly at higher simplification levels, important details are omitted. For example, in Simplification Level-3, Case-2, LCWL and LCWL+SCE+FT omit the cost of NASA's plan.

*Handling of technical terms*: The models sometimes struggle with domain-specific terms (e.g., "CTCs" in Simplification Level-4, Case-2), leaving them unexplained.

## 8 Conclusion

This paper introduced Label Confidence Weighted Learning (LCWL), a novel approach to multi-level sentence simplification that leverages weak supervision from a large paraphrase dataset and a pre-trained classifier. LCWL incorporates a label confidence weighting scheme in the training loss of the encoder-decoder model, enabling it to generate high-quality simplifications across multiple complexity levels while mitigating the impact of noisy pseudo-labels. This sets LCWL apart from existing confidence-weighting methods that primarily focus on classification tasks.

Experiments on the Newsela-auto dataset demonstrated that LCWL outperforms state-of-the-art unsupervised baselines, and when combined with fine-tuning on in-domain labeled data (LCWL+FT), it consistently delivers superior simplifications compared to strong supervised methods. The effectiveness of LCWL highlights the importance of label confidence weighting techniques for learning with noisy pseudo-labels in text simplification tasks involving encoder-decoder architectures.

Future work could explore extending LCWL to other text generation tasks facing similar challenges, investigate techniques to further improve pseudo-label quality, and conduct human evaluations to assess the perceived quality and usefulness of the generated simplifications for different target audiences.

## Limitation

While our proposed Label Confidence Weighted Learning (LCWL) approach has demonstrated promising results in multi-level sentence simplification, there are several limitations to consider:

LCWL relies on a pre-trained classifier to generate pseudo-labels for the paraphrase dataset. The quality of these pseudo-labels directly impacts the performance of the simplification model. Although the label confidence weighting scheme mitigates the impact of noisy labels, the approach is still dependent on the accuracy of the initial classifier.

Our experiments focused on English sentence simplification. Adapting LCWL to other languages would require language-specific paraphrase datasets and pre-trained classifiers, which may not be readily available for all languages. The effectiveness of LCWL in multilingual or cross-lingual settings needs further investigation.

Addressing these limitations in future work would help to further validate the effectiveness of LCWL and expand its applicability to a wider range of text simplification scenarios.

## Acknowledgments

We thank the anonymous reviewers for their helpful comments and suggestions.

## A Implementation Details

We train the classifier based on the implementation of bert-base-uncased[2] and the generator based on the implementation of bart-large[3] from Hugging Face. All experiments are conducted on a NVIDIA GeForce RTX 3080.

| Training multilevel classifier | |
|---|---|
| Hyper-Parameter | Value |
| Max source length | 256 |
| Max target length | 256 |
| Batch size | 8 |
| Learning rate | 1e-5 |
| Grad. Accu. Steps. | 1 |
| Evaluation strategy | Steps |
| Eval step | 500 |
| Optim metric | $F1_{macro}$ |
| Max steps | 6000 |

| Training multilevel generator using paraphrase | |
|---|---|
| Hyper-Parameter | Value |
| Max source length | 256 |
| Max target length | 256 |
| Batch size | 1 |
| Learning rate | 2e-5 |
| Grad. Accu. Steps. | 2 |
| Evaluation strategy | Steps |
| Eval step | 2000 |
| Optim metric | Sari |
| Max steps | 50000 |

| Finetuning multilevel generator using Newsela | |
|---|---|
| Hyper-Parameter | Value |
| Max source length | 256 |
| Max target length | 256 |
| Batch size | 1 |
| Learning rate | 2e-5 |
| Grad. Accu. Steps. | 2 |
| Evaluation strategy | Steps |
| Eval step | 1000 |
| Max steps | 40000 |

Table 6: Hyper-parameters settings for training/finetuning classifier and generator. For training multilevel generator, the batch size is set to 1 because "out of the memory". For the other hyper-parameters, we kept the default settings.

## B Multi-level Classifier Selection

For the given Newsela dataset, we finetuned the pretrained language models with different structures (bert-base-uncased, Roberta-base, and electra-base-discriminator). Additionally, we trained a simple feature-based xgboost classifier. The features we extract as described below:

- Number of Chars
- Number of Words
- Maximum Dependency Tree Depth
- Word Rank
- FKGL
- RSRS (Martinc et al., 2021)

For classifier result comparison, please see Table 7. In our final decision, we select bert-base-uncased and incorporate it with SCE. Compared with BERT only, the BERT+SCE improves the F1 from 0.444 to 0.452.

| | $F1_{macro}$ | | | | | |
|---|---|---|---|---|---|---|
| | S4 | S3 | S2 | S1 | Ori | Ave |
| Bert | 0.34 | 0.4 | 0.39 | 0.38 | 0.699 | 0.444 |
| Roberta | 0.31 | 0.37 | 0.38 | 0.35 | 0.679 | 0.417 |
| Electra | 0.39 | 0.35 | 0.37 | 0.39 | 0.68 | 0.435 |
| xgboost | 0.32 | 0.3 | 0.22 | 0.2 | 0.726 | 0.353 |
| Bert+SCE | 0.39 | 0.42 | 0.41 | 0.33 | 0.719 | **0.452** |

Table 7: The result of multi-level classifiers on Newsela test-set. The two decoupled hyperparameters, $\alpha$ and $\beta$, in SCE loss are set to 0.1 and 1 respectively in our work.

## C Multi-Level Classifier Training Set Statistics

| Total | Simp4 | Simp3 | Simp2 | Simp1 | Original |
|---|---|---|---|---|---|
| 3125 | 317 | 529 | 666 | 602 | 1011 |

Table 8: Newsela-auto multi-level classification training set statistics

The Newsela-auto training set suffers from an imbalanced distribution (as indicated in Table 8), limited scale, and label noise, as it is derived from document-level Newsela-manual using neural model extraction. Consequently, the BERT-based multi-level classifier achieves an average recall of only 0.44 on the test set, motivating our proposed methodology to mitigate classifier errors when constructing pseudo training data for text simplification

---

[2] https://huggingface.co/google-bert/bert-base-uncased

[3] https://huggingface.co/facebook/bart-large

# D Case Study

**1. Simplification Level-1:**

In Case-1, LCWL+SCE+FT, LCWL+FT and SCE+FT split the original sentence into two, making it easier to read while preserving the main ideas. LCWL doesn't split the sentence, making it slightly more complex.

In Case-2, LCWL+SCE+FT split the sentence and provide additional context about John Boehner and Mitch McConnell. LCWL+FT does not split the sentence but contextualize the person discussed. SCE+FT splits the sentence and LCWL shortens and simplifies the words.

| Method | Sentence |
| --- | --- |
| Simp Level-1 | |
| Case-1 | |
| Original | So the crowd charged the vehicles as they pulled up in clouds of choking dust, pushing and shoving through the doors, their meager possessions stuffed in cloth bundles and old suitcases. |
| Reference | So the crowd charged the vehicles as they pulled up in clouds of choking dust. People pushed and shoved through the doors, their meager possessions stuffed in cloth bundles and old suitcases. |
| LCWL | The crowd rushed the vehicles as they came up in a cloud of choking dust, pushing and shoving through the door with their meager belongings and old suitcases. |
| SCE+FT | So the crowd charged the vehicles as they pulled up in clouds of choking dust. The crowd was pushing and shoving through the doors, their meager possessions stuffed in cloth bundles and old suitcases. |
| LCWL+FT | So the crowd charged the vehicles as they pulled up in clouds of choking dust. They pushed and shoving through the doors, their meager possessions stuffed in cloth bundles and old suitcases. |
| LCWL+SCE+FT | The crowd charged the vehicles as they pulled up in clouds of choking dust, pushing and shoving through the doors. Their meager possessions were packed in cloth bundles and old suitcases. |
| Case-2 | |
| Original | However, officials said the White House was in contact throughout Thursday with some lawmakers, including House Speaker John Boehner, R-Ohio, and Senate Minority Leader Mitch McConnell, R-Ky. |
| Reference | However, officials said the White House was in contact with some lawmakers, including the Speaker of the House of Representatives, John Boehner, a Republican from Ohio, and Mitch McConnell of Kentucky, who is the leader of the Republicans in the Senate. |
| LCWL | the White House was in contact with some lawmakers throughout the day, including House Speaker Boehner and Senate Minority Leader Mitch McConnell of Kentucky. |
| SCE+FT | However, officials said the White House was in contact throughout Thursday with some lawmakers. These include House Speaker John Boehner of the House of Representatives and Senate Minority Leader Mitch McConnell of the Senate. |
| LCWL+FT | Officials said the White House was in contact throughout Thursday with some lawmakers, including House Speaker John Boehner, a Republican, and Senate Minority Leader Mitch McConnell, a Kentucky senator. |
| LCWL+SCE+FT | However, officials said the White House was in contact throughout Thursday with some lawmakers. Those lawmakers included House Speaker John Boehner of Ohio and Senate Minority Leader Mitch McConnell of Kentucky. |

## 2. Simplification Level-2:

In Case-1, LCWL+SCE+FT and LCWL+FT perform the best by splitting the original sentence and clearly stating the purpose of the scientists' study. LCWL maintains the core information but doesn't split the sentence, while SCE+FT oversimplifies by omitting key details.

In Case-2, LCWL+FT performs the best by splitting the information into three sentences and providing a clear explanation of the purpose of the flights. LCWL and LCWL+SCE+FT also perform well, but LCWL does not split. SCE+FT oversimplifies and omits crucial information.

| Method | Sentence |
| --- | --- |
| Simp Level-2 | |
| Case-1 | |
| Original | The scientists studied 22 very different species of finned swimmers, using video recordings, lab studies and computer modeling to determine what pattern, if any, might exist. |
| Reference | The scientists studied 22 very different species of finned swimmers using video recordings, lab studies and computer modeling. They wanted to find out what pattern, if any, might exist. |
| LCWL | the scientists studied 22 different species of finned swimmers using video recordings, lab studies and computer models to determine what pattern might exist. |
| SCE + FT | The scientists studied 22 very different species of finned swimmers. |
| LCWL + FT | The scientists studied 22 very different species of finned swimmers. They used video recordings, lab studies and computer modeling to determine what pattern, if any, might exist. |
| LCWL + SCE + FT | The scientists studied 22 very different species of finned swimmers. They used video recordings, lab studies and computer modeling to determine what pattern, if any, might exist. |
| Case-2 | |
| Original | Officials said there were no plans to evacuate those Americans from Iraq but that the United States was conducting enhanced intelligence flights over Irbil with both manned and unmanned aircraft to monitor the deteriorating conditions. |
| Reference | Officials said the United States was conducting flights over Irbil with both manned and unmanned aircraft. They were trying to gather information and monitor the worsening conditions. |
| LCWL | There were no plans to evacuate the Americans from Iraq but the United States was conducting enhanced intelligence flights over Irbil with both manned and unmanned aircraft to monitor the situation. |
| SCE + FT | Officials said there were no plans to evacuate those Americans from Iraq. However, that the United States was conducting enhanced intelligence flights over Irbil. |
| LCWL + FT | Officials said there were no plans to evacuate those Americans from Iraq. However, they added, the United States was conducting enhanced intelligence flights over Irbil. The flights are intended to monitor the deteriorating conditions. |
| LCWL + SCE + FT | Officials said there were no plans to evacuate those Americans from Iraq. However, they said the United States was conducting special operations over Irbil to monitor the deteriorating conditions. |

## 3. Simplification Level-3:

In Case-1, LCWL+SCE+FT and SCE+FT perform well by splitting the information and simplifying the expression. LCWL+FT splits the sentence but LCWL does not.

In Case-2, LCWL + FT perform well by splittig the sentence into three and simplifying the words. LCWL also simplifies the sentence well but omits the cost of the plan. LCWL+SCE+FT, SCE+FT and LCWL perform equally well.

| Method | Sentence |
|---|---|
| Simp Level-3 | |
| Case-1 | |
| Original | The virus has struck at a time when eggs were enjoying relative popularity in the food market; last year, the U.S. produced nearly 100 billion eggs. |
| Reference | Last week, the U.S. Department of Agriculture predicted that 95.9 billion eggs will be laid this year, while nearly 100 billion eggs were produced last year. The virus struck at a time when eggs were becoming pretty popular. |
| LCWL | the virus struck at a time when eggs were in high demand in the food market. |
| SCE+FT | The virus has struck at a time when eggs were enjoying a lot of attention. Last year, there were nearly 100 billion. |
| LCWL+FT | The virus has struck at a time when eggs are very popular. Last year, the U.S. produced nearly 100 billion eggs. |
| LCWL+SCE+FT | The virus has struck at a time when eggs were in great supply. Last year, the U.S. produced nearly 100 billion eggs. |
| Case-2 | |
| Original | While most of us hope to dodge space rocks, NASA has unveiled an ambitious, $105 million plan to build a spaceship to drag one closer to Earth. |
| Reference | Most of us hope to dodge space rocks. But NASA has an ambitious, $105 million plan. It wants to build a spaceship and drag one closer to Earth. It's a first step in a future voyage to Mars. |
| LCWL | most of us hope to avoid the space rocks, but NASA is planning to build a spaceship to bring one closer to earth. |
| SCE+FT | Most of us hope to dodge space rocks. But NASA has unveiled an ambitious, $105 million plan to build a spaceship to drag one closer to Earth. |
| LCWL+FT | Most of us hope to get to space. NASA has a plan to get rid of space rocks. It would spend $105 million to build a spaceship to drag one closer to Earth. |
| LCWL+SCE+FT | Most of us hope to avoid space rocks. NASA has unveiled a plan to build a spaceship to drag one closer to Earth. |

**4. Simplification Level-4:**

In Case-1, LCWL+SCE+FT, LCWL+FT, and SCE+FT performs the best by splitting the original sentence into two simple, easy-to-understand sentences. LCWL does not split the sentence but simplify the expression.

In Case-2, LCWL+SCE+FT and LCWL both perform well by simplifying the sentence structure and and the expression. SCE+FT splits the sentence but the expression is not simplified. LCWL simples the expression but does not split the sentence.

| Method | Sentence |
|---|---|
| Simp Level-4 | |
| Case-1 | |
| Original | The hunt won't be easy: The perfect asteroid will be tough to find because it will be too small to reflect light. |
| Reference | The hunt won't be easy. The perfect asteroid will be too small to reflect light. |
| LCWL | The hunt will not be easy because the asteroid will be too small to reflect light. |
| SCE + FT | The hunt won't be easy: The perfect asteroid will be tough to find because it will be too small to reflect light. |
| LCWL + FT | It will be hard to find the perfect asteroid. That's because it will be too small to absorb energy from the right kind of light. |
| LCWL + SCE + FT | The perfect asteroid will be hard to find. It will be too small to reflect light. |

| Case-2 | |
|---|---|
| Original | Hawking and many other physicists find CTCs abhorrent, because any macroscopic object traveling through one would inevitably create paradoxes where cause and effect break down. |
| Reference | Hawking and many other physicists don't like the idea of CTCs. Anything traveling through one would create paradoxes. |
| LCWL | Hawking and many other physicists find CTCs abhorrent because any macroscopic object moving through one would create a paradox. |
| SCE + FT | Hawking and many other physicists find CTCs uncomfortable. Any macroscopic object traveling through one would inevitably create paradoxes where cause and effect break down. |
| LCWL + FT | He and many other scientists do not agree with CTCs. They think that any object moving through one would create new ideas. |
| LCWL + SCE + FT | Hawking and many other scientists do not like CTCs. They avoid them because any large object traveling through one would create problems. |

## E  LLM Prompt Template

*Prompt template*

Please simplify the following source sentences to a simplification version of a given target level while preserving the original meaning. Target-level 4 indicates the simplest version while the Targe-level 1 the least simple version. The following are three examples of source and target sentences for each level of simplification:

Source sentence: {Original Sentence}
Target-level 1 simplification: {Target-level Simplified Sentence}
……
Source sentence: {Original Sentence}
Target-level 2 simplification: {Target-level Simplified Sentence}
……
Source sentence: {Original Sentence}
Target-level 3 simplification: {Target-level Simplified Sentence}
……
Source sentence: {Original Sentence}
Target-level 4 simplification: {Target-level Simplified Sentence}
……
Source sentence: {Original Sentence}
Target-level xx simplification: {Outputs}

Figure 2 The prompt template for multi-level sentence simplification